%% file: root.tex
\documentclass[letterpaper, 10 pt, journal, twoside]{IEEEtran}

\IEEEoverridecommandlockouts                              %

\usepackage{graphics} %
\usepackage{times} %
\usepackage{amsmath} %
\usepackage{amssymb}  %
\usepackage{balance}
\usepackage[svgnames]{xcolor}
\usepackage{colortbl}
\usepackage{url}
\usepackage{svg}
\usepackage{array}
\usepackage{multirow}
\usepackage{makecell}       %
\usepackage{cite}
\usepackage{arydshln}
\usepackage{nicematrix}
\usepackage{subcaption}
\usepackage{graphbox}
\usepackage{soul}
\usepackage{tabularx}
\usepackage{threeparttable}
\usepackage{siunitx}
\usepackage{pifont}%
\newcommand{\cmark}{\ding{51}}%
\newcommand{\xmark}{\ding{55}}%
\usepackage[accsupp]{axessibility}  %

\makeatletter
\let\NAT@parse\undefined
\makeatother

\usepackage{hyperref}
\hypersetup{
    colorlinks=true,
    linkcolor=blue,
    filecolor=magenta,
    citecolor=black,
    urlcolor=cyan,
}
\usepackage[capitalize]{cleveref} %

\crefname{section}{Sec.}{Secs.}
\Crefname{section}{Section}{Sections}
\crefname{table}{Tab.}{Tabs.}
\Crefname{table}{Table}{Tables}

\newcommand{\eg}{\emph{e.g.},}
\newcommand{\ie}{\emph{i.e.},}

\newcommand{\hotflocpp}{HOTFLoc++}

\DeclareSIUnit{\pp}{p.p.}
\DeclareSIUnit{\nothing}{\relax}

\makeatletter
\let\old@makecaption\@makecaption
\renewcommand{\@makecaption}[2]{\old@makecaption{\small #1}{\small #2}}
\makeatother

\begin{document}
\title{
HOTFLoc++: End-to-End Hierarchical LiDAR Place Recognition, Re-Ranking, and 6-DoF Metric Localisation in Forests
}

\author{Ethan Griffiths$^{1,2}$, Maryam Haghighat$^{2}$, Simon Denman$^{2}$, Clinton Fookes$^{2}$, and Milad Ramezani$^{1}$
\thanks{
Manuscript received: November, 06, 2025; Revised February, 20, 2026; Ac-
cepted April, 01, 2026.
This paper was recommended for publication by Editor H. Moon upon evaluation of the Associate Editor and Reviewers’ comments.
This work was supported by CSIRO and QUT. \emph{(Corresponding Author: Ethan Griffiths)}}
\thanks{
$^{1}$Ethan Griffiths and Milad Ramezani are with CSIRO Robotics.
E-mails: {\tt\footnotesize \emph{firstname.lastname}@data61.csiro.au}
}
\thanks{
$^{2}$Ethan Griffiths, Maryam Haghighat, Simon Denman and Clinton Fookes are with School of Electrical Engineering and Robotics, Queensland University of Technology (QUT), Brisbane, Australia.
E-mails: {\tt\footnotesize \emph\{maryam.haghighat, s.denman, c.fookes\}@qut.edu.au}
}
\thanks{Digital Object Identifier (DOI): see top of this page.}
}

\markboth{IEEE Robotics and Automation Letters. Preprint Version. Accepted April, 2026}
{Griffiths \MakeLowercase{\textit{et al.}}: HOTFLoc++: End-to-End Hierarchical Metric Localisation}

\maketitle
\input{chapters/0_abstract}


\input{chapters/1_introduction}

\input{chapters/2_relatedworks}
\input{chapters/3_method}
\input{chapters/4_experiments}
\input{chapters/5_conclusion}

\balance{}

\bibliographystyle{IEEEtran} %
\bibliography{IEEEabrv,ref}

\end{document}

%% file: chapters/0_abstract.tex
\begin{abstract}
This article presents \hotflocpp, an end-to-end hierarchical framework for LiDAR place recognition, re-ranking, and 6-DoF metric localisation in forests. Leveraging an octree-based transformer, our approach extracts features at multiple granularities to increase robustness to clutter, self-similarity, and viewpoint changes in challenging scenarios, including ground-to-ground and ground-to-aerial in forest and urban environments. We propose learnable multi-scale geometric verification to reduce re-ranking failures due to degraded single-scale correspondences. Our joint training protocol enforces multi-scale geometric consistency of the octree hierarchy via joint optimisation of place recognition with re-ranking and localisation, improving place recognition convergence. Our system achieves comparable or lower localisation errors to baselines, with runtime improvements of almost two orders of magnitude over RANSAC-based registration for dense point clouds. Experimental results on public datasets show the superiority of our approach compared to state-of-the-art methods,
achieving an average Recall@1 of 90.7\% on CS-Wild-Places: an improvement of 29.6 percentage points over baselines, while maintaining high performance on single-source benchmarks with an average Recall@1 of 91.7\% and 97.9\% on Wild-Places and MulRan, respectively. Our method achieves under 2\,m and 5$^{\circ}$ error for 97.2\% of 6-DoF registration attempts, with our multi-scale re-ranking module reducing localisation errors by $\sim\!2\times$ on average.
The code is available at \url{https://github.com/csiro-robotics/HOTFLoc}.

\end{abstract}

%% file: chapters/1_introduction.tex
\section{Introduction}
\label{sec:introduction}

Place Recognition (PR) and metric localisation are fundamental for long-term mobile robot autonomy in GPS-denied environments, yet most existing methods are tailored to structured indoor or urban environments and struggle to generalise to natural settings such as forests. These environments lack distinctive, persistent landmarks and instead exhibit strong self-similarity, clutter, and seasonal variability, undermining both keypoint-based matching and geometric verification. This is exacerbated in cross-source settings, where data is captured from distinct viewpoints or sensors (\ie{} ground \emph{vs.} aerial), causing low scene overlap and distribution shifts.

Typically, 6-DoF re-localisation in large-scale environments follows a two-step process: (1) retrieval-based PR (\ie{} coarse localisation), and (2) 6-DoF pose estimation between the query and top place candidate. To ensure successful registration, re-ranking is often employed to filter erroneous retrievals~\cite{vidanapathiranaSpectralGeometricVerification2023}. In the point cloud domain, geometric consistency has proven a strong prior for verifying retrieval quality. However, current approaches evaluate it only at a fine-grained level. This is
insufficient in complex and cross-domain scenarios where keypoints at a single scale fail to capture hierarchical structures in the data that address the inherent ambiguity of homogeneous scenes.
In cross-source settings, single-scale features are more prone to degradations than multi-scale features~\cite{guanCrossLoc3DAerialGroundCrossSource2023}, further harming robustness. Additionally, re-ranking is typically applied ad hoc to LiDAR PR, and has not been explored as an additional training signal to leverage the geometric constraints induced by re-rankers.

\begin{figure}[t]
    \centering
    \includegraphics[width=0.95\linewidth,
    trim=0 0.45cm 0 0.25cm, clip]{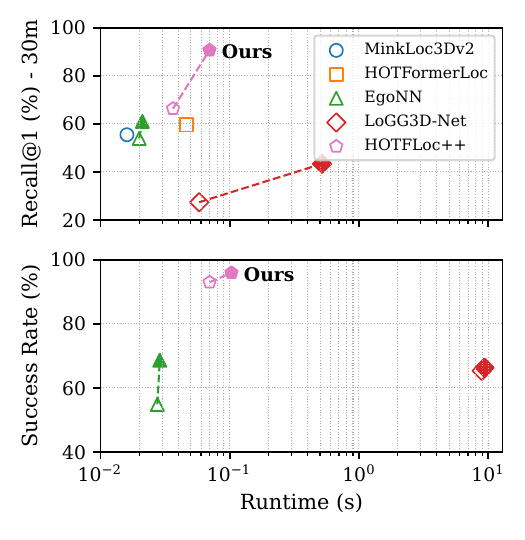}
    \caption{\hotflocpp{} achieves Pareto-optimality for place recognition (top) and metric localisation (bottom) on CS-Wild-Places. Filled symbols denote results after re-ranking.}
    \label{fig:hero}
    \vspace{-6mm}
\end{figure}

Additionally, existing approaches for 6-DoF re-localisation typically rely on keypoint detection, which struggles to produce repeatable keypoints in repetitive or cluttered environments~\cite{carvalhodelimaOnline6DoFGlobal2025}, or on robust solvers to find correspondences, which are often too slow for real-time deployment. We argue that multi-scale feature fusion is essential to maximise information extracted from such environments and to improve robustness to aliasing, occlusion, and seasonal changes.

To this end, we propose \hotflocpp{}, a unified and hierarchical end-to-end network for LiDAR PR, re-ranking, and 6-DoF metric localisation. We leverage the hierarchical features of HOTFormerLoc~\cite{griffithsHOTFormerLocHierarchicalOctree2025} to train a multi-scale geometric verification (MSGV) module that analyses geometric consistency at multiple granularities to choose the best candidate for re-localisation. Our keypoint-free re-localisation module leverages the octree hierarchy of HOTFormerLoc to process features in a coarse-to-fine manner via a patch-wise registration scheme. Importantly, we achieve low registration errors without relying on robust solvers such as RANSAC, with almost two orders of magnitude faster inference on very dense point clouds. Our robust hierarchical pipeline achieves significant performance gains on challenging natural environment datasets (\cref{fig:hero}), while retaining high performance on urban datasets, as demonstrated by our extensive experiments.
Notably, we propose a joint training protocol which optimises PR, re-ranking, and re-localisation simultaneously, maximising metric localisation performance in unstructured environments. The joint training further improves PR performance by enforcing multi-scale geometric constraints across the octree hierarchy, aiding descriptor consistency.

Our contributions include:
(a) \hotflocpp{}, an end-to-end pipeline for PR, re-ranking, and 6-DoF metric localisation, achieving an optimal performance/efficiency trade-off through joint training with complementary re-ranking and localisation objectives; 
(b) MSGV, a learnable re-ranking approach leveraging multi-scale correspondences for robustness to degraded single-scale correspondences;
and (c) Extensive experiments and comparisons with state-of-the-art baselines, demonstrating the effectiveness of the proposed framework.

%% file: chapters/2_relatedworks.tex
\section{Related Works}
\label{sec:relatedworks}

\subsection{LiDAR Place Recognition}
\label{sec:rw_lpr}

LiDAR place recognition (LPR) is typically formulated as a retrieval problem, where point clouds are encoded into descriptors then queried from a database. Prior to deep learning approaches, handcrafted descriptors~\cite{kimScanContextEgocentric2018,xuRINGRotoTranslationInvariant2023}
were common, but have since been superseded by methods trained via metric learning. These utilise three main types of feature encoder: PointNet~\cite{uyPointNetVLADDeepPoint2018,duDH3DDeepHierarchical2020},
Sparse CNNs~\cite{komorowskiMinkLoc3DPointCloud2021,cattaneoLCDNetDeepLoop2022,vidanapathiranaLoGG3DNetLocallyGuided2022,komorowskiImprovingPointCloud2022,komorowskiEgoNNEgocentricNeural2022}, and transformers~\cite{huiPyramidPointCloud2021,xuTransLoc3DPointCloud2023,goswamiSALSASwiftAdaptive2024,griffithsHOTFormerLocHierarchicalOctree2025}.
Sparse CNN methods outperformed early transformer methods in speed and accuracy~\cite{vidanapathiranaLoGG3DNetLocallyGuided2022,komorowskiEgoNNEgocentricNeural2022}, but recent approaches
have bridged this gap~\cite{goswamiSALSASwiftAdaptive2024,griffithsHOTFormerLocHierarchicalOctree2025}.

However, most LPR research has focused on urban environments, while research into unstructured natural environments has lagged. Wild-Places~\cite{knightsWildPlacesLargeScaleDataset2023} released the first large-scale dataset for long-term LPR in forests, identifying the domain gap challenging existing SOTA models. Cross-source PR has also lagged, with recent works like CrossLoc3D~\cite{guanCrossLoc3DAerialGroundCrossSource2023} exploring ground-to-aerial LPR in campus environments. HOTFormerLoc~\cite{griffithsHOTFormerLocHierarchicalOctree2025} proposed CS-Wild-Places: an aerial extension to Wild-Places, and demonstrated the effectiveness of hierarchical transformers in single- and cross-source settings.

\subsection{Metric Localisation}
\label{sec:rw_metriclocalisation}
Existing approaches for unified LPR and 6-DoF metric localisation typically fall into two categories: (a) sparse keypoint-based~\cite{duDH3DDeepHierarchical2020,komorowskiEgoNNEgocentricNeural2022}, or (b) dense correspondence-based~\cite{cattaneoLCDNetDeepLoop2022,goswamiSALSASwiftAdaptive2024}. Sparse keypoint methods such as EgoNN~\cite{komorowskiEgoNNEgocentricNeural2022} predict a keypoint saliency map to filter uncertain keypoints, with RANSAC~\cite{fischlerRandomSampleConsensus1981} for subsequent pose estimation. This works well in environments with distinct geometric features, but keypoint repeatability suffers in unstructured and cluttered environments~\cite{carvalhodelimaOnline6DoFGlobal2025}. Dense approaches typically apply robust solvers to local feature correspondences which incurs high computational cost, especially with poor initial correspondences. LCDNet~\cite{cattaneoLCDNetDeepLoop2022} employs an Unbalanced Optimal Transport (UOT) head to efficiently estimate 6-DoF pose during training, but relies on RANSAC at inference for robustness.

Recent works in point cloud registration explore more sophisticated approaches. Deep robust estimators such as PointDSC~\cite{baiPointDSCRobustPoint2021} aim to be drop-in replacements for RANSAC, training a small network with a spatial consistency prior to predict outliers. CoFiNet~\cite{yuCoFiNetReliableCoarsetofine2021} and GeoTransformer~\cite{qinGeoTransformerFastRobust2023} adopt keypoint-free coarse-to-fine registration schemes, which utilise patch-level correspondences to improve robustness in low-overlap scenes. We demonstrate that coarse-to-fine registration is better suited to unstructured environments and cross-source scenarios than keypoint-based methods.

\subsection{Re-Ranking}
\label{sec:rw_reranking}

SpectralGV (SGV)~\cite{vidanapathiranaSpectralGeometricVerification2023} pioneered geometric verification re-ranking for LPR. Leveraging a spectral technique to capture spatial consistency of correspondences, it achieves sub-linear time complexity with comparable performance to RANSAC geometric verification. However, SGV considers correspondences at a fixed feature granularity, lacking adaptiveness to degraded correspondences. We propose a learnable alternative that jointly considers the spatial consistency of multi-scale correspondences, improving robustness when a single feature resolution is not sufficient to determine correspondences. Such scenarios can occur when traversing between environments with varying geometric properties (\eg{} urban to forest), or in cross-source settings as observed in~\cite{guanCrossLoc3DAerialGroundCrossSource2023}.

%% file: chapters/3_method.tex
\begin{figure*}[t]
    \centering
    \includegraphics[width=0.95\textwidth]{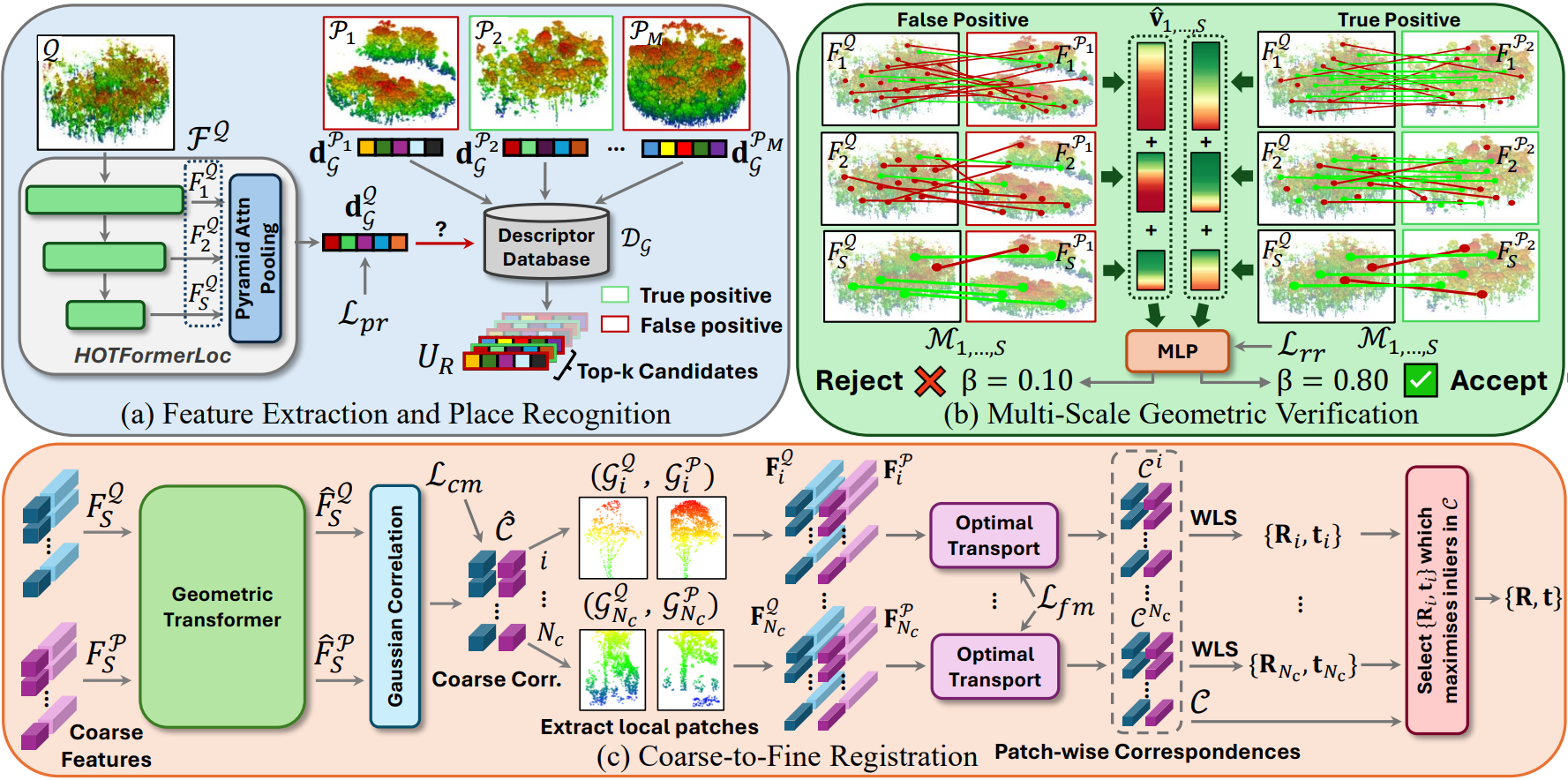}
    \caption{Pipeline of \hotflocpp{}. (a) HOTFormerLoc~\cite{griffithsHOTFormerLocHierarchicalOctree2025} extracts multi-scale local features and a robust global descriptor for PR. (b) Our learnable Multi-Scale Geometric Verification re-ranks retrievals, improving robustness to erroneous single-resolution correspondences. (c) Coarse-to-fine registration extracts patch-level correspondences and refines the patch-wise registration which maximises global inliers.}
    \label{fig:pipeline}
    \addtocounter{subfigure}{0}
    \refstepcounter{subfigure}\label{fig:pipeline:a}
    \refstepcounter{subfigure}\label{fig:pipeline:b}
    \refstepcounter{subfigure}\label{fig:pipeline:c}
    \vspace{-5mm}
\end{figure*}

\section{Method} \label{sec:method}
In this section, we detail our proposed \hotflocpp{} for end-to-end LiDAR PR (LPR), re-ranking, and 6-DoF metric localisation. Our entire pipeline is depicted in \cref{fig:pipeline}.

\subsection{Place Recognition with Hierarchical Features} \label{sec:method_placerec}
We formulate place recognition as a retrieval problem. Let $\mathcal{Q} = \{\mathbf{q}_i \in \mathbb{R}^3\}_{i=1}^{N}$ be a query point cloud with $N$ points captured by a LiDAR sensor. Let $\mathcal{D} = \{\mathcal{P}_{1},\ldots,\mathcal{P}_M \}$ be a database of $M$ point clouds captured in a prior session, where $\mathcal{P}_i = \{\mathbf{p}_j \in \mathbb{R}^3\}_{j=1}^{N_i}$ and $N_i$ is variable for each point cloud. In LPR, the goal is to retrieve point cloud $\mathcal{P}_{i} \in \mathcal{D}$ which represents the same place as $\mathcal{Q}$, ideally with as much overlap as possible. We employ HOTFormerLoc~\cite{griffithsHOTFormerLocHierarchicalOctree2025} as our place recognition backbone, which leverages an octree-based hierarchical transformer to extract strong multi-scale local features. These are pooled with a pyramid attentional pooling layer to produce a robust global descriptor. As we demonstrate in \cref{sec:experiments}, multi-scale features are essential for ensuring robustness of re-ranking and metric localisation.

Formally, our backbone encoder (\cref{fig:pipeline:a}) learns a function $f_\theta$, parametrised by $\theta$, that maps a point cloud to a set of multi-scale local features and a global descriptor
\begin{equation} \label{eq:network_mapping}
    f_\theta : \mathcal{P} \rightarrow (\mathcal{F}, \mathbf{d}_{\mathcal{G}}),
    \quad
    \mathcal{F} = [F_1,\ldots,F_S]
\end{equation}
where
$F_s = \{\mathbf{d}_{s_i} \in \mathbb{R}^{d_s}\}_{i=1}^{K_s}$
is a set of $K_s$ local descriptors of dimension $d_s$ from level $s\in\{1,\ldots,S\}$ of the feature pyramid,
which are aggregated into a $d$-dimensional global descriptor $\mathbf{d}_{\mathcal{G}} \in \mathbb{R}^d$ via pyramid attention pooling~\cite{griffithsHOTFormerLocHierarchicalOctree2025}.

At inference, the global descriptor $\mathbf{d}_\mathcal{G}^\mathcal{Q}$ of query $\mathcal{Q}$ is matched with a database of descriptors $\mathcal{D}_\mathcal{G}$ to obtain the $\text{top-}k$ retrievals $U_R =[\mathcal{P}_{R_1},\ldots,\mathcal{P}_{R_k}]$, ordered by similarity.

\subsection{Learnable Multi-Scale Geometric Verification} \label{sec:method_multiscale_reranking}
Using compact global descriptors for retrieval enables fast inference in large-scale environments, but inevitably produces false positives in ambiguous or aliased scenes. Re-ranking addresses this by analysing the local descriptors of the $\text{top-}k$ retrieval candidates in $U_R$, and re-ordering them based on a fitness score, producing $U_{RR} =[\mathcal{P}_{RR_1},\ldots,\mathcal{P}_{RR_k}]$. In particular, geometric consistency (GC) re-ranking methods such as SpectralGV (SGV), which exploit the spatial consistency of local feature correspondences, have proven to be robust and effective for LPR~\cite{vidanapathiranaSpectralGeometricVerification2023,knightsGeoAdaptSelfSupervisedTestTime2024}.
However, existing GC approaches only consider a single feature granularity, and handcrafted approaches such as~\cite{vidanapathiranaSpectralGeometricVerification2023} require further work to integrate multi-scale correspondences. We argue this limitation reduces the effectiveness of GC-based re-ranking when features of one resolution are degraded, which can occur in cross-source PR settings~\cite{guanCrossLoc3DAerialGroundCrossSource2023}.

We propose a learnable re-ranking method, coined Multi-Scale Geometric Verification (MSGV), that considers geometric consistency at multiple feature granularities (\cref{fig:pipeline:b}). Consider query point cloud $\mathcal{Q}$ and retrieval candidate $\mathcal{P} \in U_R$, with corresponding multi-scale local features $\mathcal{F}^{\mathcal{Q}}$ and $\mathcal{F}^{\mathcal{P}}$. For each $F_s^{\mathcal{Q}} \in \mathcal{F}^{\mathcal{Q}}$ and $F_s^{\mathcal{P}} \in \mathcal{F}^{\mathcal{P}}$, we process all local features with a small MLP and construct a set of putative correspondences via nearest-neighbour matching
\vspace{-1mm}
\begin{equation} \label{eq:rerank_correspondences}
    \mathcal{M}_s = \{(\mathbf{q}_s^{(i)}, \mathbf{p}_s^{(i)})\}_{i=1}^{\lambda_s}
\vspace{-1mm}
\end{equation}
where only the top-$\lambda_s$ matches are kept, and $\mathbf{q}_s^{(i)}$ and $\mathbf{p}_s^{(i)}$ denote the centroids of the matched features at level $s$. We capture the pairwise length consistency between correspondences with a geometric consistency matrix $\mathbf{M}_s \in \mathbb{R}^{\lambda_s\times\lambda_s}$, with entries $m_{i,j} \in \mathbf{M}_s$ defined as
\begin{equation} \label{eq:geometric_consistency}
    m_{i,j} \!\!\:=\!\!\: \left[ 1-\frac{\sigma_{i,j}^2}{\sigma_{d}^2} \right]_+
    \!\!\!\!\!,\; \sigma_{i,j} \!\!\:=\!\!\: \left| ||\mathbf{q}_s^{(i)}-\mathbf{q}_s^{(j)}||_2
    - ||\mathbf{p}_s^{(i)}-\mathbf{p}_s^{(j)}||_2 \right|
\end{equation}
where $[\cdot]_+=\mathrm{max}(\cdot,0)$, and $\sigma_d$ is a distance threshold controlling sensitivity to length difference.

As observed in~\cite{leordeanuSpectralTechniqueCorrespondence2005}, the values of the leading eigenvector $\mathbf{v}_s$ of $\mathbf{M}_s$ can be considered as the association of each correspondence with the main cluster of $\mathbf{M}_s$, and can thus be interpreted as inlier probabilities. This is robust in the presence of outliers as the main cluster of $\mathbf{M}_s$ is statistically formed by correct correspondences, and the likelihood of outliers forming a spatially consistent cluster is low.

To seamlessly integrate the spatial consistency of our multi-scale features into a scalar fitness, we compute $\mathbf{v}_s$ for all $s \in \{1,\ldots,S\}$ and process the concatenated vectors with an MLP
to produce the fitness score $\beta\in[0,1]$.
Inspired by GeoAdapt~\cite{knightsGeoAdaptSelfSupervisedTestTime2024}, we aid the optimisation by using $\hat{\mathbf{v}}_s$ instead of the raw leading eigenvectors, where $\hat{\mathbf{v}}_s$ is $\mathbf{v}_s$ with values sorted and min-max normalised into the range $[0,1]$.
MSGV has a complexity of $\mathcal{O}(Sk\lambda_s^2)$, which scales linearly with the number of candidates $k$, ensuring our method is scalable to more candidates.
Finally, $U_R$ is arranged in decreasing order of $\beta$ score to produce re-ranked candidates $U_{RR}$.

\subsection{Coarse-to-Fine Registration} \label{sec:method_coarsetofine_localisation}

Following PR and re-ranking success, we obtain the top-candidate point cloud $\mathcal{P}$ which overlaps query $\mathcal{Q}$. Our goal is to estimate a rigid transformation $\mathbf{T}=\{\mathbf{R},\mathbf{t}\}$ which registers $\mathcal{Q}$ and $\mathcal{P}$, with 3D rotation $\mathbf{R} \in SO(3)$ and translation $\mathbf{t} \in \mathbb{R}^3$.
Existing approaches typically estimate $\mathbf{T}$ with RANSAC~\cite{fischlerRandomSampleConsensus1981}, matching dense local features~\cite{vidanapathiranaLoGG3DNetLocallyGuided2022,goswamiSALSASwiftAdaptive2024}, or a set of sparse keypoints~\cite{komorowskiEgoNNEgocentricNeural2022,cattaneoLCDNetDeepLoop2022}.
However, computing RANSAC on thousands of local features (as is the case for dense LiDAR scans) is infeasible for real-time performance.
EgoNN avoids this issue by using a small set of keypoints, but is prone to extracting degenerate keypoints in cluttered and unstructured environments such as forests, where dense foliage and a lack of distinct landmarks harm repeatability~\cite{carvalhodelimaOnline6DoFGlobal2025}.
Furthermore, as $\mathbf{T}$ is estimated using sparse keypoints, ICP is often needed to ensure a tight registration.

By contrast, we adopt a keypoint-free coarse-to-fine registration approach that enables efficient and robust correspondence prediction with low registration errors. Importantly, we estimate 6-DoF poses via a patch-to-patch registration scheme (\cref{fig:pipeline:c}), as opposed to keypoint-to-keypoint.

Given the multi-scale local features $\mathcal{F}^\mathcal{Q}$ and $\mathcal{F}^{\mathcal{P}}$, we first enhance $F^\mathcal{Q}_S$ and $F^\mathcal{P}_S$ from the coarsest level $S$ of our feature pyramid with a Geometric Transformer~\cite{qinGeoTransformerFastRobust2023} and $L_2$ normalisation to produce $\hat{F}^\mathcal{Q}_S$ and $\hat{F}^\mathcal{P}_S$.
This lightweight network leverages geometric self-attention and cross-attention layers to explicitly encode intra-point cloud geometry and capture inter-point cloud geometric consistency, improving the transformation-invariance of the features. We then establish a set of coarse correspondences $\hat{\mathcal{C}}$ by computing a Gaussian correlation matrix
$\mathbf{G} \in \mathbb{R}^{|\hat{F}_S^\mathcal{Q}| \times{} |\hat{F}_S^\mathcal{P}|}$~\cite{qinGeoTransformerFastRobust2023} between the features.
We further perform dual-normalisation on the rows and columns of $\mathbf{G}$ to suppress ambiguous matches, and finally select the largest $N_c$ entries in $\mathbf{G}$ as our coarse correspondences $\hat{\mathcal{C}}$.

We consider each coarse correspondence
as a pair of \textit{superpoints} to be matched, allowing us to leverage the fine-grained features in HOTFormerLoc's feature hierarchy with higher frequency details, enabling more robust matching than with coarse keypoints. Within the octree of HOTFormerLoc, we expand each coarse correspondence to patch-level correspondences by assigning the local features $F_1^{\mathcal{Q}}$ and centroids $\mathcal{Q}_1$ from the finest level $1 \in \{1,\ldots,S\}$ of the feature pyramid to their octree parent nodes at level $S$
\vspace{-1mm}
\begin{equation} \label{eq:coarse_to_fine_octree}
    \mathcal{G}_i^\mathcal{Q} = \{\mathbf{q}_1^{(j)} \in \mathcal{Q}_1 \: | \: \mathrm{parent}(j) = i\},
    \: i \in \{1,\ldots,N_c\}
\vspace{-1mm}
\end{equation}
where $\mathrm{parent}(\cdot)$ maps each fine octant centroid $\mathbf{q}_1^{(j)}$ to its coarse parent node $\mathbf{q}_S^{(i)} \in \mathcal{Q}_S$ at level $S$. The corresponding fine feature matrix is denoted as $\mathbf{F}_i^\mathcal{Q} \in \mathbb{R}^{|\mathcal{G}_{i}^{\mathcal{Q}}|\times{}d_1}$, where $d_1$ is the local feature dimension at level $1$.
\Cref{eq:coarse_to_fine_octree} is repeated for point cloud $\mathcal{P}$ to produce patches
$\mathcal{G}^\mathcal{P} = \{\mathcal{G}_i^\mathcal{P}\}_{i=1}^{N_c}$, 
each with corresponding features $\mathbf{F}_i^\mathcal{P} \in \mathbb{R}^{|\mathcal{G}_i^{\mathcal{P}}|\times{}d_1}$.

For each patch correspondence
$\hat{\mathcal{C}}_i = (\mathcal{G}_{i}^\mathcal{Q}, \mathcal{G}_{i}^\mathcal{P})$,
 we initialise a cost matrix
\vspace{-0mm}
\begin{equation} \label{eq:ot_cost}
    \mathbf{C}_i = \frac{\mathbf{F}_{i}^{\mathcal{Q}}(\mathbf{F}_{i}^{\mathcal{P}})^T}{\sqrt{d_1}},\:
    \mathbf{C}_i \in \mathbb{R}^{|\mathcal{G}_{i}^{\mathcal{Q}}| \times{} |\mathcal{G}_{i}^{\mathcal{P}}|}
\vspace{-0mm}
\end{equation}
which we append with a dustbin row and column filled with learnable parameter $\alpha$ to handle unmatched points.
We process $\mathbf{C}_i$ with the learnable Sinkhorn algorithm proposed in~\cite{sarlinSuperGlueLearningFeature2020} to solve the optimal transport (OT) between patches, producing soft assignment matrix $\bar{\mathbf{Z}}_i$.
We drop the dustbin to obtain $\mathbf{Z}_i$ as the confidence matrix of correspondences between $\mathcal{G}_i^{\mathcal{Q}}$ and $\mathcal{G}_i^{\mathcal{P}}$.
To reduce the impact of erroneous correspondences, we filter out matches with confidence $z_j^i \in \mathbf{Z}_i$ less than threshold $\gamma_{z}$ and select fine correspondences $\mathcal{C}_i$ through mutual top-$k$ selection on $\mathbf{Z}_i$.
Finally, we combine the fine correspondences computed for each superpoint pair into a global set of dense correspondences $\mathcal{C} = \bigcup_{i=1}^{N_c}\mathcal{C}_i$.

To estimate $\mathbf{T}$, we adopt local-to-global registration (LGR) ~\cite{qinGeoTransformerFastRobust2023}, where a transformation candidate $\mathbf{T}_i$ is proposed for each superpoint match using its fine-grained correspondences
\vspace{-0mm}
\begin{equation} \label{eq:LGR_local}
    \mathbf{R}_i,\mathbf{t}_i = \min_{\mathbf{R},\mathbf{t}} \sum_{(\mathbf{q}_{j},\mathbf{p}_{j}) \in \mathcal{C}_i}
    z_j^i \left|\left| \mathbf{R}\cdot\mathbf{q}_{j} + \mathbf{t} - \mathbf{p}_{j} \right|\right|_2
\vspace{-0mm}
\end{equation}
which we solve in closed form with Weighted Least Squares (WLS).
Then, we select the transformation $(\mathbf{R}_i,\mathbf{t}_i)$ with the highest inlier ratio over the global dense correspondence set
\vspace{-0mm}
\begin{equation} \label{eq:LGR_global}
    \mathbf{R}, \mathbf{t} = \max_{\mathbf{R}_i,\mathbf{t}_i} \sum_{(\mathbf{q}_{j},\mathbf{p}_{j}) \in \mathcal{C}}
    \Bigl[ \bigl|\bigl| \mathbf{R}_i\cdot\mathbf{q}_{j} + \mathbf{t}_i - \mathbf{p}_{j} \bigr|\bigr|_2 < \tau_a \Bigr]
\vspace{-0mm}
\end{equation}
where $[\cdot]$ denotes the Iverson bracket, and $\tau_a$ is the inlier acceptance radius. This process is repeated $N_r$ times with the surviving inliers by iteratively solving \cref{eq:LGR_local,eq:LGR_global} to produce the final estimated transformation $\mathbf{T}$.

\subsection{Joint Training Protocol} \label{sec:method_losses}
A key advantage of our holistic approach is the joint optimisation of PR, re-ranking, and 6-DoF metric localisation. We argue these tasks are complementary, and that joint optimisation improves convergence via geometric constraints on the features extracted by the network, which we validate quantitatively in \cref{sec:exp_ablations}. To train our entire pipeline end-to-end, the overall loss is formulated as $\mathcal{L} = \mathcal{L}_{pr}+\lambda_{rr}\mathcal{L}_{rr}+\lambda_{cm}\mathcal{L}_{cm}+\lambda_{fm}\mathcal{L}_{fm}$, as detailed in the following sections.

\subsubsection*{Place Recognition}
We train our PR head in a two-stage manner. In the first stage, we disable the re-ranking and re-localisation heads and train purely on the PR task, using the Truncated Smooth Average Precision (TSAP) loss defined in~\cite{komorowskiImprovingPointCloud2022} as $\mathcal{L}_{pr}$ with a large batch size. Empirically, we find this pre-training provides a stronger initialisation for the HOTFormerLoc backbone, producing better performance.

In the second stage, we enable re-ranking and re-localisation and continue to train with PR enabled. However, we observe that using large PR batches in this stage overpowers the gradients of the other losses, leading to poor overall convergence. Instead, we reduce the PR batch size and swap the TSAP loss for a batch-hard triplet margin loss~\cite{hermansDefenseTripletLoss2017} which performs better for smaller batch sizes~\cite{komorowskiImprovingPointCloud2022}.

\subsubsection*{Re-Ranking}
To efficiently train MSGV alongside the PR head, we mine a subset of hard triplets from PR batches.
Specifically, we sort triplets by anchor and negative descriptor distance and sample the top-$N_{rr}$ hardest triplets to ensure our network learns to distinguish challenging false-positives. We employ binary cross-entropy to train the module
\vspace{-0mm}
\begin{equation} \label{eq:rerank_loss}
    \mathcal{L}_{rr} = -(y\cdot\log\beta + (1-y)\log(1-\beta))
\vspace{-0mm}
\end{equation}
where $\beta$ is the fitness score predicted by MSGV, and $y$ is $1$ for a positive pair and $0$ for negative pairs.
By including re-ranking in the optimisation, as opposed to applying it ad hoc, we enforce geometric consistency of the octree hierarchy, subsequently improving global descriptor distinctiveness.

\subsubsection*{Coarse-to-Fine Registration}
To train our coarse-to-fine registration head, we employ a second training substep after computing and backpropagating the PR and re-ranking losses.
In this substep, we sample pairs of overlapping point clouds. We utilise two loss functions to jointly optimise the quality of both coarse and fine correspondences.

For coarse correspondences, we employ an Overlap-Aware Circle Loss~\cite{qinGeoTransformerFastRobust2023}, which provides smoother gradients for optimisation than typical cross-entropy and improves low-overlap patch matching via overlap-based re-weighting.
We compute this loss in both directions from $\mathcal{Q}\rightarrow\mathcal{P}$ and $\mathcal{P}\rightarrow\mathcal{Q}$ to produce the final loss $\mathcal{L}_{cm} = (\mathcal{L}_{cm}^{\mathcal{Q}} + \mathcal{L}_{cm}^{\mathcal{P}})/2$.
 
To optimise the fine correspondences, we follow SuperGlue~\cite{sarlinSuperGlueLearningFeature2020} and apply the negative log-likelihood loss on the OT assignment matrix $\bar{\mathbf{Z}}_i$ of each superpoint correspondence $\hat{\mathcal{C}}_i \in \hat{\mathcal{C}}$, averaging over all superpoint correspondences to produce $\mathcal{L}_{fm}$. 
During training, we sample
ground-truth superpoint correspondences instead of using predicted correspondences
to ensure patches have suitable overlap.

%% file: chapters/4_experiments.tex
\section{Experiments}
\label{sec:experiments}

\begin{table}[t]
    \caption{Details of training and evaluation sets. SS and CS denote single-source and cross-source datasets, respectively.}
    \label{tab:datasets}
    \centering
    \resizebox{1.0\linewidth}{!}{
    \begin{threeparttable}
        \begin{tabular}{l c c c c}
            \hline
             \multirow{2}{*}{Dataset} & \multirow{2}{*}{Split} & \multicolumn{3}{c}{Num. Submaps} \\ & & Train & Query & Database \\ \hline
             \textit{Forest -- CS:}
             & Karawatha & 37,373 & 9,549 & 17,792 \\
             \multirow{1}{*}{CS-Wild-Places\cite{griffithsHOTFormerLocHierarchicalOctree2025}}
             & Venman & 26,807 & 6,398 & 12,383 \\
             & QCAT & --- & 753 & 369 \\
             & Samford & --- & 1,309 & 4,528 \\ \hline
             \textit{Forest -- SS:}
             & Karawatha & 13,661 & 9,642 & 9,962\tnote{$\dagger$} \\
             \multirow{1}{*}{Wild-Places~\cite{knightsWildPlacesLargeScaleDataset2023}}
             & Venman & 5,435 & 6,395 & 5,868\tnote{$\dagger$} \\ \hline
             \footnotesize{\textit{Urban -- SS:}}
             & Sejong (01/02) & 35,871 & 3,453 & 3,764 \\
             \multirow{1}{*}{MulRan~\cite{kimMulRanMultimodalRange2020}}
             & DCC (01/02) & --- & 307 & 469 \\
             & Riverside (01/02) & --- & 470 & 603 \\
            \hline
        \end{tabular}
    \begin{tablenotes}
        \item [$\dagger$] Average size of database for each sequence.
    \end{tablenotes}
    \end{threeparttable}
    }
\vspace{-4mm}
\end{table}

\subsection{Datasets and Evaluation Protocol} \label{sec:exp_datasets_and_eval_protocol}
We train and evaluate \hotflocpp{} on three datasets: CS-Wild-Places~\cite{griffithsHOTFormerLocHierarchicalOctree2025}, Wild-Places~\cite{knightsWildPlacesLargeScaleDataset2023}, and MulRan~\cite{kimMulRanMultimodalRange2020}, to demonstrate the effectiveness of our approach in a diverse range of environments and scenes. See \cref{tab:datasets} for details of the training and evaluation sets used for each dataset.

For CS-Wild-Places, we follow the training and evaluation splits proposed in HOTFormerLoc\cite{griffithsHOTFormerLocHierarchicalOctree2025}, except we downsample submaps with \qty{0.4}{\m} voxels instead of \qty{0.8}{\m} to evaluate performance for very dense submaps\footnote{Although not included due to space constraints, the reported results hold a similar pattern on CS-Wild-Places with the original \qty{0.8}{\m} voxel size.}, producing \qty{62}{\kilo\nothing} points per submap on average. For Wild-Places, we follow the original training split~\cite{knightsWildPlacesLargeScaleDataset2023}, and report the inter-sequence evaluation protocol. For MulRan, we follow the Sejong and DCC splits of SpectralGV~\cite{vidanapathiranaSpectralGeometricVerification2023}, and include Riverside 01 and 02 with submaps sampled every \qty{10}{\m}. Only Sejong is used for training and validation, allowing unseen evaluation on DCC and Riverside. In all MulRan regions, sequence 01 forms the database, and sequence 02 forms the queries.

To evaluate place recognition, we compute the similarity between the global descriptors of each query and the database and collect the top-$k$ retrieval candidates. We report the Recall@$k$ (R$k$) metric for $k\in\{1,5\}$, defined as the percentage of queries where at least one top-$k$ candidate is within a $r$-metre retrieval threshold of the query. We adopt the retrieval thresholds used in previous works, with $r=\qty{3}{\m}$ for Wild-Places, $r=\qtylist{5;20}{\metre}$ for MulRan, but for CS-Wild-Places we use $r=\qtylist{10;30}{\metre}$, with the $\qty{10}{\m}$ threshold added to capture fine-grained PR performance. For re-ranking evaluation, we re-rank the top-20 retrieval candidates and re-compute all metrics under the new ranking.

To evaluate metric localisation, we estimate the 6-DoF pose between each query and top-candidate retrieved during PR evaluation, and compare the pose estimate with ground truth.
We report three metrics: success rate (SR), which measures the percentage of queries registered within \qty{2}{\m} and \qty{5}{\degree} of the ground truth pose, as well as relative rotation error (RRE) and relative translation error (RTE), as defined in~\cite{baiPointDSCRobustPoint2021}.
We report SR under two configurations: (1) we exclude PR failures (\ie{} candidates that do not overlap the query) from the evaluation to isolate metric localisation from PR performance (denoted \emph{succ.}); and (2) we compute SR over all queries, including PR failures (denoted \emph{all}).
Ground truth poses are refined with ICP to ensure accurate ground truth.
Following SGV, we average RRE and RTE over \emph{all} localisation pairs rather than pairs within \qty{2}{\m} and \qty{5}{\degree} error to capture the true performance of the system. Importantly, we report all metrics \emph{without} ICP refining the pose estimates. 

\begin{table}[t]
    \centering
    \caption{\hotflocpp{} hyperparameters per dataset.}
    \label{tab:model_hyperparams}
    \begin{tabular}{c|c|c|c}
        \hline
        Hyperparameter & CS-Wild-Places & Wild-Places & MulRan \\
        \hline
        Learning Rate & \num{8e-4} & \num{3e-4} & \num{8e-4} \\
        Octree Attn. Type & Cartesian & Cylindrical & Cartesian \\
        $\sigma_d$ & \qty{5.0}{\metre} & \qty{1.6}{\metre} & \qty{0.4}{\metre} \\
        $\tau_a$ & \qty{1.6}{\metre} & \qty{1.6}{\metre} & \qty{0.6}{\metre} \\
        \hline
    \end{tabular}
    \vspace{-5mm}
\end{table}

\begin{table*}[t]
    \caption{Place recognition and 6-DoF metric localisation results on CS-Wild-Places~\cite{griffithsHOTFormerLocHierarchicalOctree2025} \emph{baseline} set.}
    \label{tab:cswp_baseline_results}
    \centering
    \resizebox{1.0\textwidth}{!}{
    \begin{threeparttable}
        \begin{tabular}{l|c|cc|cc|ccc|cc|cc|ccc}
            \hline
            \multirow{3}{*}{Method} & \multirow{3}{*}{Re-Ranker} & \multicolumn{7}{c|}{Karawatha} & \multicolumn{7}{c}{Venman}  \\ \cline{3-16}
            & & \multicolumn{2}{c|}{PR (10~m)} & \multicolumn{2}{c|}{PR (30~m)} & \multicolumn{3}{c|}{6-DoF Metric Localisation} & \multicolumn{2}{c|}{PR (10~m)} & \multicolumn{2}{c|}{PR (30~m)}  & \multicolumn{3}{c}{6-DoF Metric Localisation}\\ %
            & & R1 & R5 & R1 & R5 & SR (succ. / all) & RTE [m] & RRE [$^\circ$] & R1 & R5 & R1 & R5  & SR (succ. / all) & RTE [m] & RRE [$^\circ$] \\
            \hline
            MinkLoc3Dv2~\cite{komorowskiImprovingPointCloud2022} & --- & 31.1 & 54.2 & 56.1 & 70.2 & --- & --- & --- & 33.8 & \underline{63.6} & \underline{61.2} & \underline{79.2} & --- & --- & --- \\
            EgoNN~\cite{komorowskiEgoNNEgocentricNeural2022} & --- & 34.3 & 61.4 & \underline{60.0} & \underline{76.2} & 43.3\% / 23.0\% & 7.81 & 44.64 & \underline{37.7} & 62.3 & 59.8 & 77.4 & 31.7\% / 15.2\% & 11.20 & 63.69 \\
            LoGG3D-Net\tnote{$*\ddagger$}~~~\cite{vidanapathiranaLoGG3DNetLocallyGuided2022} & --- & 18.7 & 37.7 & 33.7 & 50.8 & \underline{78.3\%} / \underline{24.7\%} & \underline{3.39} & \underline{11.70} & 11.7 & 29.6 & 26.1 & 48.1 & \underline{95.1\%} / \underline{25.2\%} & \underline{1.24} & \underline{3.75} \\
            HOTFormerLoc~\cite{griffithsHOTFormerLocHierarchicalOctree2025} & --- & \underline{34.4} & \underline{61.5} & 57.0 & 72.1 & --- & --- & --- & 29.9 & 57.0 & 47.7 & 69.4 & --- & --- & ---   \\ %
            \textbf{\hotflocpp{} (Ours)} & --- & \textbf{44.4} & \textbf{71.2} & \textbf{72.0} & \textbf{84.1} & \textbf{82.9\%} / \textbf{59.1\%} & \textbf{2.32} & \textbf{5.74} & \textbf{38.4} & \textbf{68.1} & \textbf{65.2} & \textbf{82.1} & \textbf{96.9\%} / \textbf{63.6\%} & \textbf{0.58} & \textbf{1.47}   \\
            \hline
            EgoNN~\cite{komorowskiEgoNNEgocentricNeural2022} & SGV~\cite{vidanapathiranaSpectralGeometricVerification2023} & 48.6 & 68.3 & 64.6 & 79.7 & 56.9\% / 36.7\% & 4.70 & 26.79 & 33.9 & 60.4 & 52.5 & 79.7 & 40.1\% / 22.8\% & 9.61 & 51.57 \\
            LoGG3D-Net\tnote{$*\ddagger$}~~~\cite{vidanapathiranaLoGG3DNetLocallyGuided2022} & SGV~\cite{vidanapathiranaSpectralGeometricVerification2023} & 28.9 & 45.1 & 45.8 & 55.3 & 81.5\% / 35.6\% & 2.50 & 9.18 & 34.1 & 52.4 & 64.4 & 66.7 & 97.9\% / 63.1\% & 0.55 & 1.99 \\
            \textbf{\hotflocpp{} (Ours)} & SGV~\cite{vidanapathiranaSpectralGeometricVerification2023} & \textbf{67.5} & \textbf{82.4} & \textbf{85.9} & \textbf{89.0} & \underline{85.9\%} / \textbf{70.6\%} & \underline{1.70} & \underline{3.37} & \underline{70.9} & \underline{84.1} & \underline{91.0} & \underline{92.5} & \underline{98.7\%} / \underline{88.9\%} & \underline{0.42} & \underline{1.00} \\
            \textbf{\hotflocpp{} (Ours)} & \textbf{MSGV (Ours)} & \underline{66.1} & \underline{81.1} & \underline{81.6} & \underline{88.2} & \textbf{88.9\%} / \underline{70.1\%} & \textbf{1.11} & \textbf{1.57} & \textbf{77.8} & \textbf{86.1} & \textbf{93.5} & \textbf{93.9} & \textbf{99.0\%} / \textbf{91.6\%} & \textbf{0.38} & \textbf{0.85} \\
            \hline
        \end{tabular}
        \begin{tablenotes}
        \item [$*$] Method uses 1024-dimensional global descriptors, instead of 256-dimensional. \hspace{45mm}$^\ddagger$\,Method uses 0.8\,m voxelised data instead of 0.4\,m to remain tractable.
        \end{tablenotes}
    \end{threeparttable}
    }
    \vspace{-1mm}
\end{table*}

\begin{table*}[t]
    \caption{Place recognition and 6-DoF metric localisation results on CS-Wild-Places~\cite{griffithsHOTFormerLocHierarchicalOctree2025} \emph{unseen} set.}
    \label{tab:cswp_unseen_results}
    \centering
    \resizebox{1.0\linewidth}{!}{
    \begin{threeparttable}
        \begin{tabular}{l|c|cc|cc|ccc|cc|cc|ccc}
            \hline
            \multirow{3}{*}{Method} & \multirow{3}{*}{Re-Ranker} & \multicolumn{7}{c|}{QCAT} & \multicolumn{7}{c}{Samford}  \\ \cline{3-16}
            & & \multicolumn{2}{c|}{PR (10~m)} & \multicolumn{2}{c|}{PR (30~m)} & \multicolumn{3}{c|}{6-DoF Metric Localisation} & \multicolumn{2}{c|}{PR (10~m)} & \multicolumn{2}{c|}{PR (30~m)}  & \multicolumn{3}{c}{6-DoF Metric Localisation}\\ %
            & & R1 & R5 & R1 & R5 & SR (succ. / all) & RTE [m] & RRE [$^\circ$] & R1 & R5 & R1 & R5  & SR (succ. / all) & RTE [m] & RRE [$^\circ$] \\
            \hline
            MinkLoc3Dv2~\cite{komorowskiImprovingPointCloud2022} & --- & 13.0 & \underline{43.6} & \underline{51.0} & \textbf{79.9} & --- & --- & --- & 28.2 & 53.7 & 53.6 & 71.9 & --- & --- & ---  \\
            EgoNN~\cite{komorowskiEgoNNEgocentricNeural2022} & --- & 13.5 & 37.8 & 49.4 & 67.3 & \underline{54.6\%} / \underline{23.4\%} & \underline{12.41} & 50.07 & 27.4 & 51.6 & 46.5 & 65.5 & \underline{89.8\%} / \underline{41.1\%} & \underline{2.75} & \underline{10.56}  \\
            LoGG3D-Net\tnote{$*\ddagger$}~~~\cite{vidanapathiranaLoGG3DNetLocallyGuided2022} & --- & 14.3 & 31.2 & 37.5 & 61.3 & 52.8\% / 15.4\% & 15.33 & \underline{33.09} & 3.4 & 9.6 & 12.3 & 23.5 & 35.1\% / 3.0\% & 22.12 & 44.33 \\
            HOTFormerLoc~\cite{griffithsHOTFormerLocHierarchicalOctree2025} & --- & \underline{14.7} & 42.9 & \textbf{52.5} & \underline{74.1} & --- & --- & --- & \underline{37.5} & \underline{66.7} & \underline{63.0} & \underline{77.6} & --- & --- & ---   \\ %
            \textbf{\hotflocpp{} (Ours)} & --- & \textbf{19.7} & \textbf{44.5} & 46.7 & 66.0 & \textbf{95.8\%} / \textbf{51.3\%} & \textbf{1.31} & \textbf{2.40} & \textbf{48.4} & \textbf{79.4} & \textbf{81.4} & \textbf{89.5} & \textbf{96.3\%} / \textbf{79.6\%} & \textbf{1.56} & \textbf{1.93}  \\
            \hline
            EgoNN~\cite{komorowskiEgoNNEgocentricNeural2022} & SGV~\cite{vidanapathiranaSpectralGeometricVerification2023} & 40.0 & 54.6 & 53.7 & 71.6 & 79.3\% / 40.1\% & 4.86 & 22.31 & 63.9 & 68.4 & 73.4 & 76.6 & \underline{98.3\%} / 70.6\% & \underline{1.24} & 1.79  \\
            LoGG3D-Net\tnote{$*\ddagger$}~~~\cite{vidanapathiranaLoGG3DNetLocallyGuided2022} & SGV~\cite{vidanapathiranaSpectralGeometricVerification2023} & 18.5 & 36.5 & 42.5 & 64.6 & 47.2\% / 18.0\% & 12.61 & 31.51 & 6.2 & 13.3 & 20.8 & 28.2 & 38.6\% / 8.9\% & 20.72 & 32.69 \\
            \textbf{\hotflocpp{} (Ours)} & SGV~\cite{vidanapathiranaSpectralGeometricVerification2023} & \underline{61.1} & \underline{68.8} & \underline{71.2} & \underline{79.8} & \textbf{98.2\%} / \underline{70.4\%} & \textbf{0.51} & \textbf{0.78} & \underline{65.4} & \underline{90.5} & \underline{94.5} & \underline{95.2} & 96.8\% / \underline{90.0\%} & 1.32 & \underline{1.47}  \\
            \textbf{\hotflocpp{} (Ours)} & \textbf{MSGV (Ours)} & \textbf{72.8} & \textbf{83.5} & \textbf{92.2} & \textbf{95.0} & \underline{96.7\%} / \textbf{88.8\%} & \underline{0.77} & \underline{1.55} & \textbf{71.6} & \textbf{94.1} & \textbf{95.3} & \textbf{96.5} & \textbf{99.1\%} / \textbf{93.8\%} & \textbf{1.17} & \textbf{0.97} \\
            \hline
        \end{tabular}
        \begin{tablenotes}
        \item [$*$] Method uses 1024-dimensional global descriptors, instead of 256-dimensional. \hspace{45mm}$^\ddagger$\,Method uses 0.8\,m voxelised data instead of 0.4\,m to remain tractable.
        \end{tablenotes}
    \end{threeparttable}
    }
    \vspace{-4mm}
\end{table*}

\subsection{Implementation Details} \label{sec:exp_implementation_details}
See \cref{tab:model_hyperparams} for dataset-specific hyperparameters.
We train \hotflocpp{} on a single NVIDIA H100 GPU. 
\hotflocpp{} employs a lightweight version of the HOTFormerLoc~\cite{griffithsHOTFormerLocHierarchicalOctree2025} backbone with $S=3$ pyramid levels and a maximum channel size of 192. Our network has \qty{14.8}{\mega\nothing} parameters in total, of which \qty{1.6}{\mega\nothing} belong to the re-ranking and metric localisation heads. During training, we apply data augmentations including random point jitter, random point removal, random translations within $\pm\,$\qty{5}{\m}, random rotations about the z-axis between $\pm\,180^\circ$, and random occlusions up to \qty{45}{\degree}.

In our two-stage training, we pre-train the PR head with batch size 2048 for 60 epochs, followed by 60 epochs with batch size 256 and losses $\mathcal{L}_{rr}$, $\mathcal{L}_{cm}$, and $\mathcal{L}_{fm}$ enabled, where $\lambda_{rr}\!=\!\lambda_{cm}\!=\!\lambda_{fm}\!=\!1$. In both stages, we reduce the learning rate by a factor of 10 after 40 epochs, and apply a memory-efficient sharpness-aware loss~\cite{duSharpnessAwareTrainingFree2022} after 10 epochs to encourage convergence to a flat minima. In MSGV, we sample $\lambda_s\in\{512,256,128\}$ correspondences from fine to coarse levels, and approximate leading eigenvectors via the power method for 5 iterations. In our metric localisation head, we set the number of top-$k$ coarse correspondences to $N_c = 256$, and point confidence threshold to $\gamma_z = 0.05$.

\subsection{Results}
\subsubsection*{CS-Wild-Places}
We compare our method with end-to-end LPR methods including MinkLoc3Dv2~\cite{komorowskiImprovingPointCloud2022}, EgoNN~\cite{komorowskiEgoNNEgocentricNeural2022}, LoGG3D-Net~\cite{vidanapathiranaLoGG3DNetLocallyGuided2022}, and HOTFormerLoc~\cite{griffithsHOTFormerLocHierarchicalOctree2025}. All methods produce 256-dimensional global descriptors, except for LoGG3D-Net as we adopt the variant with 1024-dimensional descriptors used in~\cite{vidanapathiranaSpectralGeometricVerification2023,knightsWildPlacesLargeScaleDataset2023}. We adapt LoGG3D-Net for metric localisation via RANSAC feature matching, mirroring the approach used in~\cite{vidanapathiranaSpectralGeometricVerification2023}. We compare our MSGV re-ranking with SpectralGV~\cite{vidanapathiranaSpectralGeometricVerification2023}. \Cref{tab:cswp_baseline_results,tab:cswp_unseen_results} show results for the \emph{baseline} and \emph{unseen} splits of CS-Wild-Places.

\hotflocpp{} excels in the challenging cross-source ground-to-aerial re-localisation setting, outperforming previous methods by a significant margin with and without re-ranking enabled. Without re-ranking, our backbone achieves a
9.4 and 12.4 percentage point~(p.p.) average improvement in Recall@1 over EgoNN for the \qty{10}{\m} and \qty{30}{\m} retrieval thresholds, respectively. With SGV re-ranking enabled, this improvement increases to 
19.6~p.p. and 24.6~p.p., and with MSGV it further rises to 25.5~p.p. and 29.6~p.p. over EgoNN. 

Notably, our MSGV re-ranking outperforms SGV, with a significant Recall@1 improvement of up to 21.0~p.p. on QCAT. Empirically, we find that QCAT exhibits significant perceptual aliasing, causing a large number of false positive correspondences at the feature resolution used by SGV. Our method filters out these false positives by jointly considering the geometric consistency of correspondences at different granularities, improving robustness in the presence of degraded single-resolution correspondences. Additionally, MSGV identifies higher overlap candidates for registration, reducing metric localisation error by $\sim\!2\times$ on average.

The 6-DoF metric localisation results highlight the weaknesses of keypoint-based methods in dense forests, with \hotflocpp{} achieving an average SR of
95.9\% on PR successes and 86.1\% on all queries with re-ranking enabled, compared to EgoNN's average SR of 68.7\% on PR successes and 42.6\% on all queries. This difference is largely due to the challenges of producing repeatable keypoints in dense forests~\cite{carvalhodelimaOnline6DoFGlobal2025}. The cross-source nature of CS-Wild-Places also plays a significant role, as the varying densities of points captured from the ground and aerial perspectives biases keypoints towards regions that may not be well sampled from the other perspective. By contrast, the coarse-to-fine registration of \hotflocpp{} considers patch-to-patch matches, which can still produce accurate registrations under low overlap. Furthermore, our method's robust hypothesise-and-verify approach prevents erroneous patch correspondences from corrupting the registration. Our 6-DoF metric localisation approach also consistently outperforms the RANSAC-based registration used in LoGG3D-Net, with significantly better generalisation to the unseen forests, whilst requiring two orders of magnitude less runtime (\cref{tab:runtime_analysis}).

\begin{table*}[t]
    \caption{Place recognition and 6-DoF metric localisation results on Wild-Places~\cite{knightsWildPlacesLargeScaleDataset2023} inter-sequence protocol.}
    \label{tab:wp_results}
    \centering
    \resizebox{1.00\linewidth}{!}{
    \begin{threeparttable}
        \begin{tabular}{l|c|cc|ccc|cc|ccc}
            \hline
            \multirow{3}{*}{Method} & \multirow{3}{*}{Re-Ranker} & \multicolumn{5}{c|}{Karawatha} & \multicolumn{5}{c}{Venman} \\ \cline{3-12}
            & & \multicolumn{2}{c|}{PR (3m)} & \multicolumn{3}{c|}{6-DoF Metric Localisation} & \multicolumn{2}{c|}{PR (3m)} & \multicolumn{3}{c}{6-DoF Metric Localisation} \\
            & & R1 & R5 & SR (succ. / all) & RTE [m] & RRE [$^\circ$] & R1 & R5 & SR (succ. / all) & RTE [m] & RRE [$^\circ$] \\
            \hline
            MinkLoc3Dv2~\cite{komorowskiImprovingPointCloud2022} & --- & 67.8 & 92.6 & --- & --- & --- & 75.8 & 96.1 & --- & --- & --- \\
            EgoNN~\cite{komorowskiEgoNNEgocentricNeural2022} & --- & 70.9 & \underline{92.8} & 67.7\% / 67.2\% & 1.14 & 10.53 & 77.2 & 95.6 & 77.3\% / 76.8\% & \underline{0.90} & 7.91 \\
            LoGG3D-Net\tnote{*}~~\cite{vidanapathiranaLoGG3DNetLocallyGuided2022} & --- & \textbf{74.7} & 92.4 & \textbf{96.3\%} / \textbf{95.1\%} & \textbf{0.37} & \textbf{2.34} & 79.8 & 93.6 & \textbf{98.0\%} / \textbf{96.1\%} & \textbf{0.52} & \textbf{2.03} \\
            HOTFormerLoc~\cite{griffithsHOTFormerLocHierarchicalOctree2025} & --- &  69.6 & 92.2 & --- & --- & --- & \textbf{80.1} & \underline{96.4} & --- & --- & --- \\
            \textbf{\hotflocpp{} (Ours)} & --- & \underline{71.1} & \textbf{93.4} & \underline{89.2\%} / \underline{88.9\%} & \underline{0.88} & \underline{4.53} & \underline{79.9} & \textbf{96.9} & \underline{95.7\%} / \underline{95.7\%} & \textbf{0.52} & \underline{2.66}  \\
            \hline
            EgoNN~\cite{komorowskiEgoNNEgocentricNeural2022} & SGV~\cite{vidanapathiranaSpectralGeometricVerification2023} & 90.1 & 97.5 & 74.0\% / 73.6\% & 0.49 & 5.67 & \underline{96.6} & \underline{99.3} & 82.2\% / 81.9\% & 0.37 & 3.90 \\
            LoGG3D-Net\tnote{*}~~\cite{vidanapathiranaLoGG3DNetLocallyGuided2022} & SGV~\cite{vidanapathiranaSpectralGeometricVerification2023} & \underline{91.6} & 97.3 & \textbf{97.9\%} / \textbf{97.5\%} & \textbf{0.24} & \textbf{1.06} & \textbf{97.0} & 98.4 & \textbf{99.9\%} / \textbf{99.7\%} & \textbf{0.17} & \textbf{0.59} \\
            \textbf{\hotflocpp{} (Ours)} & SGV~\cite{vidanapathiranaSpectralGeometricVerification2023} & \textbf{91.7} & \textbf{98.0} & \underline{96.6\%} / \underline{96.6\%} & \underline{0.38} & 2.49 & 95.9 & \textbf{99.6} & 99.4\% / 99.4\% & 0.29 & 1.57 \\
            \textbf{\hotflocpp{} (Ours)} & \textbf{MSGV (Ours)} & 89.4 & \underline{97.8} & \underline{96.6\%} / 96.3\% & 0.39 & \underline{2.39} & 94.0 & \textbf{99.6} & \underline{99.6\%} / \underline{99.6\%} & \underline{0.28} & \underline{1.28} \\ %
            \hline
        \end{tabular}
        \begin{tablenotes} %
        \item [$*$] Method uses 1024-dimensional global descriptors, instead of 256-dimensional.
        \end{tablenotes}
    \end{threeparttable}
    }
    \vspace{-1mm}
\end{table*}

\begin{table*}[t]
    \caption{Place recognition and 6-DoF metric localisation results on MulRan~\cite{kimMulRanMultimodalRange2020}.}
    \label{tab:mulran_results}
    \centering
    \resizebox{1.0\linewidth}{!}{
    \begin{threeparttable}
        \begin{tabular}{l|cc|cc|ccc|cc|cc|ccc|cc|cc|ccc}
            \hline
            \multirow{3}{*}{Method} & \multicolumn{7}{c|}{Sejong 02} & \multicolumn{7}{c|}{DCC 02}  & \multicolumn{7}{c}{Riverside 02} \\ \cline{2-22} %
            & \multicolumn{2}{c|}{PR (5~m)} & \multicolumn{2}{c|}{PR (20~m)} & \multicolumn{3}{c|}{6-DoF Metric Loc.} & \multicolumn{2}{c|}{PR (5~m)} & \multicolumn{2}{c|}{PR (20~m)}  & \multicolumn{3}{c|}{6-DoF Metric Loc.} & \multicolumn{2}{c|}{PR (5~m)} & \multicolumn{2}{c|}{PR (20~m)} & \multicolumn{3}{c}{6-DoF Metric Loc.} \\ %
            & R1 & R5 & R1 & R5 & SR (succ. / all) & RTE & RRE & R1 & R5 & R1 & R5  & SR (succ. / all) & RTE & RRE & R1 & R5 & R1 & R5  & SR (succ. / all) & RTE & RRE \\
            \hline
            EgoNN~\cite{komorowskiEgoNNEgocentricNeural2022} & \textbf{98.2} & \textbf{99.7} & \textbf{99.2} & \textbf{99.8} & \underline{99.8\%} / \textbf{99.1\%} & \underline{0.20} & \underline{0.40} & 68.1 & 86.6 & 89.9 & 94.5 & \underline{97.1\%} / 87.0\% & 0.62 & \underline{0.89} & \textbf{72.9} & 86.2 & 84.1 & 91.0 & \underline{98.4\%} / \underline{81.7\%} & \textbf{0.27} & \underline{0.48} \\
            LoGG3D-Net\tnote{*}~~\cite{vidanapathiranaLoGG3DNetLocallyGuided2022} & \underline{97.3} & \textbf{99.7} & 98.6 & \textbf{99.8} & \textbf{99.9\%} / \underline{98.3\%} & \textbf{0.18} & \textbf{0.36} & \underline{69.1} & 88.9 & \underline{91.5} & 95.4 & \textbf{98.6\%} / \underline{90.2\%} & \textbf{0.25} & \textbf{0.40} & \underline{69.2} & 89.7 & 85.6 & 93.3 & \textbf{98.5\%} / 68.4\% & \underline{0.32} & \textbf{0.42} \\
            HOTFormerLoc~\cite{griffithsHOTFormerLocHierarchicalOctree2025} & 96.9 & \underline{99.3} & \textbf{99.2} & \underline{99.7} & --- & --- & --- & \textbf{69.4} & 93.2 & \textbf{95.4} & 97.1 & --- & --- & --- & 66.0 & \underline{91.0} & 88.2 & 95.3 & --- & --- & --- \\
            \textbf{\hotflocpp{}} & 96.8 & 99.2 & \underline{99.0} & 99.5 & 97.0\% / 95.8\% & 0.43 & 1.15 & 67.8 & \textbf{94.5} & \textbf{95.4} & \textbf{98.4} & 88.7\% / 84.4\% & 1.20 & 3.68 & 67.3 & 90.5 & \underline{89.9} & \underline{96.3} & 88.5\% / 79.4\% & 1.09 & 2.37  \\
            \textbf{\hotflocpp{}\tnote{$\dagger$}} & 97.0 & \underline{99.3} & \textbf{99.2} & \underline{99.7} & 98.8\% / 97.9\% & 0.23 & \textbf{0.36} & 68.1 & \underline{93.5} & \textbf{95.4} & \underline{97.7} & 96.6\% / \textbf{91.2\%} & \underline{0.42} & 1.12 & \underline{69.2} & \textbf{92.7} & \textbf{92.3} & \textbf{97.0} & 96.2\% / \textbf{88.6\%} & 0.56 & 1.09 \\
            \hline
            EgoNN~\cite{komorowskiEgoNNEgocentricNeural2022}~(SGV) & \underline{97.0} & 99.0 & 98.2 & \textbf{99.9} & \underline{99.7\%} / 97.9\% & \underline{0.19} & 0.39 & \textbf{74.6} & 93.2 & 97.4 & 97.7 & \textbf{99.3\%} / \underline{96.1\%} & \textbf{0.18} & 0.55 & \underline{80.0} & 90.5 & 87.1 & 93.8 & \textbf{99.5\%} / 86.5\% & \textbf{0.20} & 0.48  \\
            LoGG3D-Net\tnote{*}~~\cite{vidanapathiranaLoGG3DNetLocallyGuided2022} (SGV) & \textbf{98.8} & \textbf{99.9} & 99.4 & \textbf{99.9} & \textbf{99.9\%} / \textbf{99.3\%} & \textbf{0.17} & 0.35 & \underline{73.0} & 91.2 & 96.7 & 96.7 & \textbf{99.3\%} / 95.8\% & \underline{0.20} & \underline{0.39} & \textbf{89.2} & \underline{96.6} & 95.1 & 98.5 & \underline{99.2\%} / \textbf{94.4\%} & \underline{0.22} & \underline{0.42} \\
            \textbf{\hotflocpp{}}~(SGV) & 94.2 & 99.6 & \textbf{99.6} & \textbf{99.9} & 96.9\% / 96.3\% & 0.39 & 0.83 & 71.0 & \textbf{96.1} & \textbf{99.3} & \textbf{99.7} & 91.8\% / 88.3\% & 0.96 & 4.95 & 76.3 & 95.3 & \underline{95.7} & 97.6 & 89.3\% / 84.1\% & 0.96 & 2.48  \\
            \textbf{\hotflocpp{}\tnote{$\dagger$}}~~(SGV) & 95.1 & \underline{99.8} & \underline{99.5} & \textbf{99.9} & 98.9\% / \underline{98.3\%} & \underline{0.19} & \textbf{0.26} & 71.0 & \underline{95.8} & \underline{98.7} & \textbf{99.7} & \underline{98.0\%} / \textbf{96.7\%} & 0.22 & \textbf{0.36} & 79.6 & \textbf{97.4} & \textbf{96.8} & \textbf{99.1} & 97.0\% / 93.8\% & 0.45 & 0.54  \\
            \textbf{\hotflocpp{}}~(\textbf{MSGV}) & 93.2 & 99.5 & 99.2 & \underline{99.8} & 95.3\% / 95.0\% & 0.46 & 0.74 & 68.1 & \underline{95.8} & 97.7 & \underline{99.0} & 90.4\% / 87.3\% & 1.19 & 4.24 & 75.5 & 95.3 & \textbf{96.8} & 98.3 & 88.5\% / 82.4\% & 0.96 & 2.55 \\
            \textbf{\hotflocpp{}}\tnote{$\dagger$}~~(\textbf{MSGV}) & 94.4 & 99.5 & 99.2 & \underline{99.8} & 98.1\% / 97.2\% & 0.28 & \underline{0.31} & 72.3 & 94.1 & 96.7 & 97.7 & 97.2\% / 92.5\% & 0.28 & 1.57 & 77.0 & 96.3 & \textbf{96.8} & \underline{98.7} & 97.7\% / \underline{94.2\%} & 0.26 & \textbf{0.37} \\ %
            \hline
        \end{tabular}
        \begin{tablenotes}
        \item [$*$] Method uses 1024-dimensional global descriptors, instead of 256-dimensional.\hspace{133mm}$^\dagger$\,4-layer version of \hotflocpp{}.
        \end{tablenotes}
    \end{threeparttable}
    }
    \vspace{-4mm}
\end{table*}

\subsubsection*{Wild-Places}
We evaluate our approach on the Wild-Places dataset in \cref{tab:wp_results}. For place recognition, our approach consistently reports higher Recall@5 than baselines, achieving 97.8\% and 99.6\% on Karawatha and Venman, respectively.
LoGG3D-Net maintains the highest Recall@1 across both splits, but requires a bulkier 1024-dimensional global descriptor to achieve this. Our MSGV re-ranking improves Recall@1 by up to
21.7~p.p. and 39.6~p.p. on Karawatha and Venman, respectively.
We observe SGV achieves 2.1 p.p. higher Recall@1 on average with our backbone, but we argue this small trade-off on single-source data is justified for the improved robustness seen in cross-source environments.

For 6-DoF metric localisation, LoGG3D-Net achieves the best average SR before and after re-ranking. This is not unexpected as it relies on RANSAC feature matching between dense local features, which takes \qty{873}{\ms} on average for Wild-Places submaps. In contrast, our \hotflocpp{} achieves comparable RTE and RRE with a total runtime of \qty{101}{\milli\second} on the same hardware, with only \qty{34.7}{\milli\second} of that time required for metric localisation. See \cref{sec:exp_runtime_analysis} for further runtime comparisons. Compared with keypoint-based re-localisation, the advantages of \hotflocpp{} are evident, achieving an average SR of
98.0\%: a 20.2~p.p. improvement over EgoNN.

\begin{table}[t]
    \caption{Runtime analysis on CS-Wild-Places~\cite{griffithsHOTFormerLocHierarchicalOctree2025}.}
    \label{tab:runtime_analysis}
    \centering
    \resizebox{1.0\linewidth}{!}{
    \begin{threeparttable}
        \begin{tabular}{l|cc|cc|cc|c}
            \hline
            \multirow{2}{*}{Method} & \multicolumn{2}{c|}{Feat. Extract.} & \multicolumn{2}{c|}{Re-Ranking} & \multicolumn{2}{c|}{Metric Loc.} & \multirow{2}{*}{\makecell{Total [ms]}}  \\
            & mean & std & mean & std & mean & std & \\

            \hline
            EgoNN (SGV) & \textbf{19.9} & \underline{4.4} & \textbf{1.2} & \textbf{1.3} & \textbf{7.6} & \underline{1.0} & \textbf{28.7} \\
            LoGG3D-Net\tnote{$\ddagger$}~~(SGV) & \underline{30.3} & \textbf{1.4} & 597.8 & 228.6 & 5955.0 & 2896.0 & 6583.1 \\
            \textbf{\hotflocpp{}} (SGV) & 36.4 & 5.8 & \underline{22.7} & \underline{2.4} & \underline{33.5} & \textbf{0.3} & \underline{92.6} \\
            \textbf{\hotflocpp{}} \textbf{(MSGV)} & 36.4 & 5.8 & 33.3 & \textbf{1.3} & \underline{33.5} & \textbf{0.3} & 103.2 \\
            \hline
            
        \end{tabular}
        
        \begin{tablenotes}
        \item [$\ddagger$] Method uses 0.8\,m voxelised data instead of 0.4\,m to remain tractable.
        \end{tablenotes}
    \end{threeparttable}
    }
    \vspace{-4mm}
\end{table}

\subsubsection*{MulRan}

We report results on the MulRan dataset in \cref{tab:mulran_results}. Note we also report results for a deeper version of our network with $S=4$ HOTFormer levels (denoted \hotflocpp{}$^\dagger$), as we empirically find that MulRan's urban setting benefits from coarser features than in forests. For LoGG3D-Net, we utilise the pre-trained weights from~\cite{vidanapathiranaSpectralGeometricVerification2023}.

While our primary focus is on unstructured forest environments, we demonstrate comparable performance with existing methods in the urban environments of MulRan. Consistent with the findings in Wild-Places, our \hotflocpp{} achieves the highest overall Recall@5 with an average of 96.9\% and 99.0\% on the \qtylist{5;20}{\metre} retrieval thresholds, respectively.
Our method also achieves high Recall@1 for the \qty{20}{\metre} retrieval threshold, with an average of 94.8\% and 97.9\% before and after re-ranking, respectively. Futhermore, our \hotflocpp{}$^\dagger$ model achieves comparable metric localisation performance to EgoNN and LoGG3D-Net, with an average SR of 97.7\% on PR successes and 94.6\% on all queries.

Interestingly, on the saturated Sejong 02 sequence, both EgoNN and our method exhibit \emph{worse} Recall@1 for the \qty{5}{\metre} threshold after re-ranking with both SGV and MSGV. Upon investigation, these failures primarily occur in a specific region with high feature ambiguity at all scales, where both EgoNN and \hotflocpp{} identify clusters of geometrically consistent but \emph{incorrect} correspondences, which is enough to compromise geometric consistency-based re-rankers. Developing re-ranking solutions that can better handle such ambiguities is a potential direction for future research. 

\begin{table}[t]
    \caption{Impact of joint training on place recognition.}
    \label{tab:ablation_joint_training}
    \centering
    \resizebox{1.0\linewidth}{!}{
    \begin{threeparttable}
    \begin{tabular}{cccc|ccc}
        \hline
        $\mathcal{L}_{pr}$ & $\mathcal{L}_{rr}$ & $\mathcal{L}_{cm}$ & $\mathcal{L}_{fm}$ & CS-Wild-Places & Wild-Places & MulRan \\
        \hline
        \cmark & \xmark & \xmark & \xmark & 31.3 & 72.9 & 82.1 \\
        \cmark & \cmark & \xmark & \xmark & 32.0 & 72.0 & 82.3 \\
        \cmark & \xmark & \cmark & \cmark & \underline{37.7} & \underline{73.6} & \underline{82.4} \\
        \cmark & \cmark & \cmark & \cmark & \textbf{38.4} & \textbf{75.5} & \textbf{83.3} \\
        \hline
    \end{tabular}
    \ \ Values indicate mean R1 for smallest retrieval threshold \emph{without} re-ranking.
    \end{threeparttable}
    }
    \vspace{-4.8mm}
\end{table}

\subsection{Runtime Analysis} \label{sec:exp_runtime_analysis}
We conduct a runtime analysis of 6-DoF re-localisation methods in \cref{tab:runtime_analysis}. Our hardware setup uses a NVIDIA H100 GPU and Intel 8452Y CPU. EgoNN achieves the fastest runtime of \qty{28.7}{\milli\second}, attributed to its lightweight CNN backbone and sparse keypoints. Whilst efficient, this approach struggles in forest environments (\cref{tab:cswp_baseline_results,tab:cswp_unseen_results,tab:wp_results}). \hotflocpp{} achieves a runtime of \qty{103.2}{\milli\second}, allowing for $\sim\,$\qty{10}{\hertz} operation.
Our metric localisation head runs over two orders of magnitude faster than LoGG3D-Net with SGV re-ranking due to the inefficiency of point-level RANSAC. While more advanced localisation methods could be integrated with LoGG3D-Net, this is outside the scope of this work. Overall, we believe our method enables online global localisation on mobile compute unites such as edge devices, which is a focus for future work.

\subsection{Ablation Study}
\label{sec:exp_ablations}

\subsubsection*{Joint Training}
In \cref{tab:ablation_joint_training} we evaluate the impact that our joint training protocol has on the PR performance of \hotflocpp{}, \emph{without} re-ranking enabled at inference. A key finding is that our re-ranking and metric localisation losses have a positive impact on PR performance. The largest increase is brought by $\mathcal{L}_{cm}$ and $\mathcal{L}_{fm}$ with an average Recall@1 improvement of 2.5~p.p., while enabling all losses improves Recall@1 by 3.6~p.p.
Indeed, these losses provide beneficial constraints on the distribution of local features during training, as both losses guide the network to extract distinctive yet geometrically consistent features which are invariant to rigid transformations and thus easier to match. This finding matches the observations in~\cite{vidanapathiranaLoGG3DNetLocallyGuided2022} that consistent local features tend to improve global descriptor repeatability.

\subsubsection*{Multi-Scale Geometric Verification}
We assess the effect of using multiple feature scales with MSGV in \cref{tab:ablation_msgv}. Across all datasets, using at least 2 scales improves performance compared to single-scale correspondences. Our method achieves the best performance on CS-Wild-Places and MulRan when using all 3 feature granularities, with Recall@1 improvements of 4.3~p.p. and 5.1~p.p., respectively.

\begin{table}[t]
    \caption{Ablation of num. feature scales used in MSGV.}
    \label{tab:ablation_msgv}
    \centering
    \begin{threeparttable}
    \begin{tabular}{l|ccc}
        \hline
        Num. Scales & CS-Wild-Places & Wild-Places & MulRan \\
        \hline
        1 (fine) & 67.8 & 90.6 & 75.6 \\
        2 (fine+mid) & \underline{69.2} & \textbf{91.7} & \underline{77.8} \\
        3 (fine+mid+coarse) & \textbf{72.1} & \underline{91.2} & \textbf{80.7} \\
        \hline
    \end{tabular}
    \ \ Values indicate mean R1 for smallest retrieval threshold.
    \end{threeparttable}
    \vspace{-3mm}
\end{table}

%% file: chapters/5_conclusion.tex
\section{Conclusion}
\label{sec:conclusion}

This paper introduces \hotflocpp{}, a unified framework trained end-to-end for LiDAR place recognition, re-ranking, and 6-DoF metric localisation. We propose a learnable multi-scale geometric verification module that improves robustness in the presence of degraded single-resolution correspondences, demonstrating significant improvements in cross-source forest environments. Our framework presents a coarse-to-fine registration approach that achieves comparable performance to RANSAC-based approaches with runtime improvements up to two orders of magnitude. Furthermore, our experiments demonstrate the complementary nature of our joint training approach, with re-ranking and metric localisation objectives contributing to higher place recognition performance. In future work, we will incorporate multi-modality to further improve robustness in challenging environments.

%% file: IEEEabrv.bib
@STRING{IEEE_J_PAMI       = "{IEEE} Trans. Pattern Anal. Machine Intell."}

@STRING{IEEE_J_RO         = "{IEEE} Trans. Robot."}

@STRING{IEEE_L_RA         = "{IEEE} Robot. Automat. Lett."}

@STRING{IEEE_C_RO         = "Proc. {IEEE/RSJ} Int. Conf. Intell. Robots Syst."}

@STRING{IEEE_C_RA         = "Proc. {IEEE} Int. Conf. Robot. Automat."}

@STRING{IEEE_C_ICCV         = "Proc. {IEEE/CVF} Int. Conf. Comput. Vis."}

@STRING{IEEE_C_CVPR         = "Proc. {IEEE/CVF} Conf. Comput. Vis. Pattern Recognit."}

@STRING{IEEE_C_NIPS         = "Proc. Adv. Neural Inf. Process. Syst."}


%% file: ref.bib
@inproceedings{baiPointDSCRobustPoint2021,
  title = {{{PointDSC}}: {{Robust Point Cloud Registration}} Using {{Deep Spatial Consistency}}},
  shorttitle = {{{PointDSC}}},
  booktitle = IEEE_C_CVPR,
  author = {Bai, Xuyang and Luo, Zixin and Zhou, Lei and Chen, Hongkai and Li, Lei and Hu, Zeyu and Fu, Hongbo and Tai, Chiew-Lan},
  year = {2021},
  month = jun,
  pages = {15854--15864},
  doi = {10.1109/CVPR46437.2021.01560},
  urldate = {2025-02-05},
  copyright = {https://ieeexplore.ieee.org/Xplorehelp/downloads/license-information/IEEE.html},
  isbn = {978-1-6654-4509-2},
  langid = {english}
}

@article{duSharpnessAwareTrainingFree2022,
  title = {Sharpness-{{Aware Training}} for {{Free}}},
  author = {Du, Jiawei and Zhou, Daquan and Feng, Jiashi and Tan, Vincent and Zhou, Joey Tianyi},
  year = {2022},
  month = dec,
  journal = IEEE_C_NIPS}

@article{goswamiSALSASwiftAdaptive2024,
  title = {{{SALSA}}: {{Swift Adaptive Lightweight Self-Attention}} for {{Enhanced LiDAR Place Recognition}}},
  shorttitle = {{{SALSA}}},
  author = {Goswami, Raktim Gautam and Patel, Naman and Krishnamurthy, Prashanth and Khorrami, Farshad},
  year = {2024},
  month = oct,
  journal = {IEEE Robot. Autom. Lett.},
  volume = {9},
  number = {10},
  pages = {8242--8249},
  issn = {2377-3766, 2377-3774},
  doi = {10.1109/LRA.2024.3440098},
  urldate = {2024-11-13},
  copyright = {https://ieeexplore.ieee.org/Xplorehelp/downloads/license-information/IEEE.html}
}

@article{guanCrossLoc3DAerialGroundCrossSource2023,
  title = {{{CrossLoc3D}}: {{Aerial-Ground Cross-Source 3D Place Recognition}}},
  shorttitle = {{{CrossLoc3D}}},
  author = {Guan, Tianrui and Muthuselvam, Aswath and Hoover, Montana and Wang, Xijun and Liang, Jing and Sathyamoorthy, Adarsh Jagan and Conover, Damon and Manocha, Dinesh},
  year = {2023},
  journal = IEEE_C_ICCV,
  pages = {11301--11310},
  publisher = {IEEE},
  address = {Paris, France},
  doi = {10.1109/ICCV51070.2023.01041},
  urldate = {2024-11-13},
  copyright = {https://doi.org/10.15223/policy-029},
  isbn = {9798350307184}
}

@inproceedings{huiPyramidPointCloud2021,
  title = {Pyramid {{Point Cloud Transformer}} for {{Large-Scale Place Recognition}}},
  booktitle = IEEE_C_ICCV,
  author = {Hui, Le and Yang, Hang and Cheng, Mingmei and Xie, Jin and Yang, Jian},
  year = {2021},
  pages = {6098--6107},
  urldate = {2023-11-28},
  langid = {english}
}

@inproceedings{kimMulRanMultimodalRange2020,
  title = {{{MulRan}}: {{Multimodal Range Dataset}} for {{Urban Place Recognition}}},
  shorttitle = {{{MulRan}}},
  booktitle = IEEE_C_RA,
  author = {Kim, Giseop and Park, Yeong Sang and Cho, Younghun and Jeong, Jinyong and Kim, Ayoung},
  year = {2020},
  pages = {6246--6253},
  issn = {2577-087X},
  doi = {10.1109/ICRA40945.2020.9197298},
  urldate = {2023-11-24}
}

@inproceedings{kimScanContextEgocentric2018,
  title = {Scan {{Context}}: {{Egocentric Spatial Descriptor}} for {{Place Recognition Within 3D Point Cloud Map}}},
  shorttitle = {Scan {{Context}}},
  booktitle = IEEE_C_RO,
  author = {Kim, Giseop and Kim, Ayoung},
  pages = {4802--4809},
  year = {2018},
  issn = {2153-0866},
  doi = {10.1109/IROS.2018.8593953},
  urldate = {2023-11-28}
}

@article{knightsGeoAdaptSelfSupervisedTestTime2024,
  title = {{{GeoAdapt}}: {{Self-Supervised Test-Time Adaptation}} in {{LiDAR Place Recognition Using Geometric Priors}}},
  shorttitle = {{{GeoAdapt}}},
  author = {Knights, Joshua and Hausler, Stephen and Sridharan, Sridha and Fookes, Clinton and Moghadam, Peyman},
  year = {2024},
  month = jan,
  journal = IEEE_L_RA,
  volume = {9},
  number = {1},
  pages = {915--922},
  issn = {2377-3766},
  doi = {10.1109/LRA.2023.3337698},
  urldate = {2024-09-09}
}

@inproceedings{knightsWildPlacesLargeScaleDataset2023,
  title = {Wild-{{Places}}: {{A Large-Scale Dataset}} for {{Lidar Place Recognition}} in {{Unstructured Natural Environments}}},
  shorttitle = {Wild-{{Places}}},
  booktitle = IEEE_C_RA,
  author = {Knights, Joshua and Vidanapathirana, Kavisha and Ramezani, Milad and Sridharan, Sridha and Fookes, Clinton and Moghadam, Peyman},
  year = {2023},
  pages = {11322--11328},
  doi = {10.1109/ICRA48891.2023.10160432},
  urldate = {2023-11-24}
}

@article{komorowskiEgoNNEgocentricNeural2022,
  title = {{{EgoNN}}: {{Egocentric Neural Network}} for {{Point Cloud Based 6DoF Relocalization}} at the {{City Scale}}},
  shorttitle = {{{EgoNN}}},
  author = {Komorowski, Jacek and Wysoczanska, Monika and Trzcinski, Tomasz},
  year = {2022},
  month = apr,
  journal = IEEE_L_RA,
  volume = {7},
  number = {2},
  pages = {722--729},
  issn = {2377-3766},
  doi = {10.1109/LRA.2021.3133593},
  urldate = {2024-08-14}
}

@inproceedings{komorowskiImprovingPointCloud2022,
  title = {Improving {{Point Cloud Based Place Recognition}} with {{Ranking-based Loss}} and {{Large Batch Training}}},
  booktitle = {26th Int. Conf. Pattern Recognit.},
  author = {Komorowski, Jacek},
  year = {2022},
  pages = {3699--3705},
  publisher = {IEEE},
  doi = {10.1109/ICPR56361.2022.9956458}
}

@inproceedings{komorowskiMinkLoc3DPointCloud2021,
  title = {{{MinkLoc3D}}: {{Point Cloud Based Large-Scale Place Recognition}}},
  shorttitle = {{{MinkLoc3D}}},
  author = {Komorowski, Jacek},
  year = {2021},
  month = jan,
  booktitle = {Proc. IEEE Winter Conf. Appl. Comput. Vis.},
  pages = {1789--1798},
  doi = {10.1109/WACV48630.2021.00183},
  urldate = {2024-11-13},
  copyright = {https://ieeexplore.ieee.org/Xplorehelp/downloads/license-information/IEEE.html},
  isbn = {9781665404778}
}

@inproceedings{leordeanuSpectralTechniqueCorrespondence2005,
  title = {A Spectral Technique for Correspondence Problems Using Pairwise Constraints},
  booktitle = {Proc. 10th {IEEE} Int. Conf. Comput. Vis.},
  author = {Leordeanu, M. and Hebert, M.},
  year = {2005},
  month = oct,
  volume = {2},
  pages = {1482-1489},
  issn = {2380-7504},
  doi = {10.1109/ICCV.2005.20},
  urldate = {2025-04-02}
}

@article{qinGeoTransformerFastRobust2023,
  title = {{{GeoTransformer}}: {{Fast}} and {{Robust Point Cloud Registration With Geometric Transformer}}},
  shorttitle = {{{GeoTransformer}}},
  author = {Qin, Zheng and Yu, Hao and Wang, Changjian and Guo, Yulan and Peng, Yuxing and Ilic, Slobodan and Hu, Dewen and Xu, Kai},
  year = {2023},
  month = aug,
  journal = IEEE_J_PAMI,
  volume = {45},
  number = {8},
  pages = {9806--9821},
  issn = {1939-3539},
  doi = {10.1109/TPAMI.2023.3259038},
  urldate = {2025-04-01}
}

@inproceedings{sarlinSuperGlueLearningFeature2020,
  title = {{{SuperGlue}}: {{Learning Feature Matching With Graph Neural Networks}}},
  shorttitle = {{{SuperGlue}}},
  booktitle = IEEE_C_CVPR,
  author = {Sarlin, Paul-Edouard and DeTone, Daniel and Malisiewicz, Tomasz and Rabinovich, Andrew},
  year = {2020},
  month = jun,
  pages = {4937--4946},
  doi = {10.1109/CVPR42600.2020.00499},
  urldate = {2025-02-05},
  copyright = {https://ieeexplore.ieee.org/Xplorehelp/downloads/license-information/IEEE.html},
  isbn = {978-1-7281-7168-5},
  langid = {english}
}

@inproceedings{uyPointNetVLADDeepPoint2018,
  title = {{{PointNetVLAD}}: {{Deep Point Cloud Based Retrieval}} for {{Large-Scale Place Recognition}}},
  shorttitle = {{{PointNetVLAD}}},
  booktitle = IEEE_C_CVPR,
  author = {Uy, Mikaela Angelina and Lee, Gim Hee},
  year = {2018},
  month = jun,
  pages = {4470--4479},
  doi = {10.1109/CVPR.2018.00470},
  urldate = {2023-11-16},
  isbn = {978-1-5386-6420-9},
  langid = {english}
}

@inproceedings{vidanapathiranaLoGG3DNetLocallyGuided2022,
  title = {{{LoGG3D-Net}}: {{Locally Guided Global Descriptor Learning}} for {{3D Place Recognition}}},
  shorttitle = {{{LoGG3D-Net}}},
  booktitle = IEEE_C_RA,
  author = {Vidanapathirana, Kavisha and Ramezani, Milad and Moghadam, Peyman and Sridharan, Sridha and Fookes, Clinton},
  pages = {2215--2221},
  year = {2022},
  doi = {10.1109/ICRA46639.2022.9811753},
  urldate = {2024-05-08}
}

@article{vidanapathiranaSpectralGeometricVerification2023,
  title = {Spectral {{Geometric Verification}}: {{Re-Ranking Point Cloud Retrieval}} for {{Metric Localization}}},
  shorttitle = {Spectral {{Geometric Verification}}},
  author = {Vidanapathirana, Kavisha and Moghadam, Peyman and Sridharan, Sridha and Fookes, Clinton},
  year = {2023},
  month = may,
  journal = IEEE_L_RA,
  volume = {8},
  number = {5},
  pages = {2494--2501},
  issn = {2377-3766},
  doi = {10.1109/LRA.2023.3255560},
  urldate = {2025-04-01}
}

@article{xuRINGRotoTranslationInvariant2023,
  title = {{{RING}}++: {{Roto-Translation Invariant Gram}} for {{Global Localization}} on a {{Sparse Scan Map}}},
  shorttitle = {{{RING}}++},
  author = {Xu, Xuecheng and Lu, Sha and Wu, Jun and Lu, Haojian and Zhu, Qiuguo and Liao, Yiyi and Xiong, Rong and Wang, Yue},
  year = {2023},
  month = dec,
  journal = IEEE_J_RO,
  volume = {39},
  number = {6},
  pages = {4616--4635},
  issn = {1941-0468},
  doi = {10.1109/TRO.2023.3303035},
  urldate = {2024-02-12}
}

@article{xuTransLoc3DPointCloud2023,
  title = {{{TransLoc3D}}: Point Cloud Based Large-Scale Place Recognition Using Adaptive Receptive Fields},
  shorttitle = {{{TransLoc3D}}},
  author = {Xu, Tian-Xing and Guo, Yuan-Chen and Li, Zhiqiang and Yu, Ge and Lai, Yu-Kun and Zhang, Song-Hai},
  year = {2023},
  journal = {Commun. Inf. Syst.},
  volume = {23},
  number = {1},
  pages = {57--83},
  issn = {15267555, 21634548},
  doi = {10.4310/CIS.2023.v23.n1.a3},
  urldate = {2024-11-13},
  langid = {english}
}

@inproceedings{yuCoFiNetReliableCoarsetofine2021,
  title = {{{CoFiNet}}: {{Reliable Coarse-to-fine Correspondences}} for {{Robust PointCloud Registration}}},
  shorttitle = {{{CoFiNet}}},
  booktitle = IEEE_C_NIPS,
  author = {Yu, Hao and Li, Fu and Saleh, Mahdi and Busam, Benjamin and Ilic, Slobodan},
  year = {2021},
  volume = {34},
  pages = {23872--23884},
  urldate = {2025-02-05}
}

@article{cattaneoLCDNetDeepLoop2022,
  title = {{{LCDNet}}: {{Deep Loop Closure Detection}} and {{Point Cloud Registration}} for {{LiDAR SLAM}}},
  shorttitle = {{{LCDNet}}},
  author = {Cattaneo, Daniele and Vaghi, Matteo and Valada, Abhinav},
  year = {2022},
  month = aug,
  journal = IEEE_J_RO,
  volume = {38},
  number = {4},
  pages = {2074--2093},
  issn = {1941-0468},
  doi = {10.1109/TRO.2022.3150683},
  urldate = {2025-09-24}
}

@inproceedings{duDH3DDeepHierarchical2020,
  title = {{{DH3D}}: {{Deep Hierarchical 3D Descriptors}} for {{Robust Large-Scale 6DoF Relocalization}}},
  shorttitle = {{{DH3D}}},
  booktitle = {Proc. Eur. Conf. Comput. Vis.},
  author = {Du, Juan and Wang, Rui and Cremers, Daniel},
  year = {2020},
  pages = {744--762},
  doi = {10.1007/978-3-030-58548-8_43},
  isbn = {978-3-030-58548-8},
  langid = {english}
}

@inproceedings{griffithsHOTFormerLocHierarchicalOctree2025,
  title = {{{HOTFormerLoc}}: {{Hierarchical Octree Transformer}} for {{Versatile Lidar Place Recognition Across Ground}} and {{Aerial Views}}},
  shorttitle = {{{HOTFormerLoc}}},
  booktitle = IEEE_C_CVPR,
  author = {Griffiths, Ethan and Haghighat, Maryam and Denman, Simon and Fookes, Clinton and Ramezani, Milad},
  year = {2025},
  pages = {6648--6658},
  langid = {english}
}

@article{fischlerRandomSampleConsensus1981,
  title = {Random Sample Consensus: A Paradigm for Model Fitting with Applications to Image Analysis and Automated Cartography},
  shorttitle = {Random Sample Consensus},
  author = {Fischler, Martin A. and Bolles, Robert C.},
  year = {1981},
  month = jun,
  journal = {Commun. ACM},
  volume = {24},
  number = {6},
  pages = {381--395},
  issn = {0001-0782},
  doi = {10.1145/358669.358692},
  urldate = {2025-10-01}
}

@inproceedings{carvalhodelimaOnline6DoFGlobal2025,
  title = {Online {{6DoF Global Localisation}} in {{Forests}} Using {{Semantically-Guided Re-Localisation}} and {{Cross-View Factor-Graph Optimisation}}},
  booktitle = IEEE_C_RO,
  year = {2025},
  author = {{Carvalho de Lima}, Lucas and Griffiths, Ethan and Haghighat, Maryam and Denman, Simon and Fookes, Clinton and Borges, Paulo and Brunig, Michael and Ramezani, Milad},
  urldate = {2025-10-01},
  langid = {english}
}

@misc{hermansDefenseTripletLoss2017,
  title = {In {{Defense}} of the {{Triplet Loss}} for {{Person Re-Identification}}},
  author = {Hermans, Alexander and Beyer, Lucas and Leibe, Bastian},
  year = {2017},
  month = nov,
  number = {arXiv:1703.07737},
  note = {arXiv:1703.07737 [cs]},
  eprint = {1703.07737},
  primaryclass = {cs},
  publisher = {arXiv},
  doi = {10.48550/arXiv.1703.07737},
  urldate = {2025-10-15},
  archiveprefix = {arXiv}
}
